\documentclass[sigconf]{aamas}

\setcopyright{none}
\copyrightyear{2024}

\acmConference[ALA '24]
    {Proc.\@ of the Adaptive and Learning Agents Workshop (ALA 2024)}
    {May 6 -- 7, 2024}
    {Online, \url{https://ala2024.github.io}}
    {Avalos, Milec, M\"uller, Wang, Yates (eds.)}
\acmYear{2024}
\acmDOI{}
\acmPrice{}
\acmISBN{}
\title{Learning to Beat ByteRL: Exploitability of Collectible Card Game Agents}

\author{Radovan Halu\v{s}ka}
\affiliation{
    \institution{Charles University}
    \city{Prague}
    \country{Czech Republic}
}
\email{radovanhaluska@duck.com}

\author{Martin Schmid}
\affiliation{
    \institution{Charles University \& EquiLibre Technologies}
    \city{Prague}
    \country{Czech Republic}
}
\email{schmidm@kam.mff.cuni.cz}

\begin{abstract}
    While Poker, as a family of games, has been studied extensively in the last decades,
    collectible card games have seen relatively little attention. Only recently have we seen an
    agent that can compete with professional human players in Hearthstone, one of the most popular
    collectible card games. Although artificial agents must be able to work with imperfect
    information in both of these genres, collectible card games pose another set of distinct
    challenges. Unlike in many poker variants, agents must deal with state space so vast that even
    enumerating all states consistent with the agent's beliefs is intractable, rendering the
    current search methods unusable and requiring the agents to opt for other techniques. In this
    paper, we investigate the strength of such techniques for this class of games. Namely, we
    present preliminary analysis results of ByteRL, the state-of-the-art agent in Legends of Code
    and Magic and Hearthstone. Although ByteRL beat a top-10 Hearthstone player from China, we show
    that its play in Legends of Code and Magic is highly exploitable.
\end{abstract}


\begin{document}

\pagestyle{fancy}
\fancyhead{}

\maketitle

\section{Introduction}
\label{section:introduction}

Games have been at the forefront of artificial intelligence research since the very
beginning~\cite{Shannon1950Chess}. They provide a well-defined set of rules for players to follow,
yet they are complex enough to pose a challenge for humans and computer agents alike. Furthermore,
they allow efficient software implementations that are currently crucial for pushing the boundaries
of artificial intelligence. The field of artificial intelligence has seen many successful attempts
at beating humans in board games, such as Checkers~\cite{Schaeffer2005Checkers},
Chess~\cite{Campbell2002DeepBlue}, Go~\cite{Silver2016MasteringGo, Silver2017MasteringChess},
Shogi~\cite{Silver2017MasteringChess} and many others. These games have one thing in common: they
are games of perfect information, meaning that both players have access to complete information
about the game at any time. In the real world, we rarely have access to the whole picture. More
often than not, we are working with partial information. Games with this characteristic feature
have seen fewer successful attempts at beating humans than their perfect information counterparts.
The most notable successes are beating humans in Limit Heads-up Texas Hold'em
Poker~\cite{Bowling2015HeadsUp}, No-Limit Heads-up Texas Hold'em Poker~\cite{Moravck2017DeepStack,
Brown2018Libratus}, Stratego~\cite{Prolat2022MasteringStratego} and
Diplomacy~\cite{Bakhtin2022Diplomacy}.

A \textit{collectible card game} (CCG) is an imperfect information card game that mixes two stages:
a deck-building or a drafting stage and a battle stage. In the drafting stage, each player builds a
deck of cards unknown to the opponent, which is then used against their opponent in the battle
stage. The goal of the game is to decrease the opponent's health to zero. There are many popular
collectible card games, such as Magic: The Gathering~\cite{Magic}, Hearthstone~\cite{Hearthstone},
The Elder Scrolls: Legends~\cite{TheElderScrolls} and many others. A trait that makes collectible
card games appealing to human players and challenging for AI agents is the broad range of ways to
mix and match available cards into decks. Even small collectible card games with tens of available
cards can offer more potential decks than the total number of atoms in the
universe~\cite{Kowalski2020ActiveGenes}. Despite the popularity of
Hearthstone,\footnote{https://activeplayer.io/hearthstone} which receives substantially more
traffic than the world's largest Poker site (PokerStars), collectible card games received very
little attention from AI researchers in previous years.

This paper looks at a collectible card game named \textit{Legends of Code and Magic}
(LOCM)~\cite{Kowalski2023SummarizingLOCM} and its state-of-the-art agent
\textit{ByteRL}~\cite{Xi2023MasteringLOCM}. Legends of Code and Magic is a game that was
specifically created to promote AI research in collectible card games. It is loosely inspired by
The Elder Scrolls: Legends~\cite{Kowalski2023SummarizingLOCM}, but it is much smaller than the most
popular CCGs, allowing quick iteration of ideas in the research phase. Even though LOCM makes a
couple of simplifying assumptions, such as no randomness in the cards, researching and training
successful agents in LOCM translates to successful agents in Hearthstone, as shown by the team at
ByteDance~\cite{Xi2023MasteringLOCM, Xiao2023MasteringHS}. We show that ByteRL is easily
exploitable in specific cases by training an adversarial agent using behaviour cloning and
subsequently fine-tuning it with the help of reinforcement learning.

In the rest of the paper, we look at the preliminary results of our experiments in which we search
for the best response against ByteRL. Section~\ref{section:background} briefly introduces
reinforcement learning and behaviour cloning, and it describes Legends of Code and Magic in greater
detail. Section~\ref{section:related-work} provides an overview of the literature and approaches
that have been applied to LOCM and other CCGs, focusing primarily on ByteRL.
Section~\ref{section:behaviour-cloning} describes the setup of the first part of our experiments
and comments on the results achieved so far. Section~\ref{section:rl-fine-tuning} continues with
the second part of our experiments that focus on RL training. Lastly,
Section~\ref{section:conclusion} concludes the paper with final remarks and future plans.

%%%%%%%%%%%%%%%%%%%%%%%%%%%%%%%%%%%%%%%%%%%%%%%%%%%%%%%%%%%%%%%%%%%%%%%%%%%%%%%%%%%%%%%%%%%%%%%%%%%

\section{Background}
\label{section:background}

This section briefly describes reinforcement learning and behaviour cloning and provides pointers
to the literature for a more in-depth treatment of the topics. The rest of the section introduces
Legends of Code and Magic. It describes the two available versions, 1.2 and 1.5, and discusses
their differences. Finally, it dives deeper into LOCM 1.5, which is the main focus of this paper.

\subsection{Reinforcement Learning}
\label{subsection:reinforcement-learning}

We consider partially observable sequential decision tasks modelled by Partially Observable Markov
Decision Processes. Partially Observable Markov Decision Process is a six-tuple $(S, A, P, R,
\Omega, O)$, where $S$ is the set of states, $A$ is the set of actions, $P: S \times A \times S \to
\mathbb{R}$ describes conditional transition probabilities, $R: S \times A \to \mathbb{R}$ is the
reward function, $\Omega$ is the set of observations, and $O: S \times A \times \Omega \to
\mathbb{R}$ describes the probability of an observation. At each timestep $t$, an agent is in state
$s_t \in S$, it observes observation $o_t \in \Omega$, and based on this observation, it selects
action $a_t \in A$. After executing action $a_t \in A$, it transitions into a new state $s_{t+1}
\in S$ according to $P$, receives a new observation $o_{t+1} \in \Omega$ according to $O$, and
finally it receives a reward $r_{t+1} \in \mathbb{R}$ according to $R$. This process repeats for $t
= 1 \dots T$, where $T$ is the maximum length of an episode.

An agent acting in an environment follows a policy $\pi(\cdot | s_t)$ that returns a probability of
taking each valid action in state $s_t$. The goal of the agent is to find an optimal policy $\pi^*$
that maximizes the expected return $G_t = \sum_{k=t}^{T} r_k$.

Agent's interaction with the environment can be summarized as a trajectory. Internally, a
trajectory is a sequence of states and actions; however, from the perspective of an agent acting in
the environment, a trajectory is a sequence of observations and actions, where each state is a
complete snapshot of the environment at a specific moment, each observation is a snapshot of the
environment available to the agent and each action is the agent's response to the observation it
received. A more in-depth treatment of the topics presented here is given in the book by Sutton and
Barto~\cite{Sutton1998RL}.

We model Legends of Code and Magic, a two-player zero-sum game, as a POMDP by fixing the opponent
and thus viewing it as part of the environment. Fixing the opponent makes the environment
single-agent from our perspective. Finding the optimal policy in such an environment corresponds to
the best response against the opponent in the original
environment~\cite{Greenwald2017BestResponse}.

\subsection{Behaviour Cloning}
\label{subsection:behaviour-cloning}

Behaviour cloning is the most straightforward approach to imitation learning, which falls under the
category of offline reinforcement learning. In offline RL, we do not interact with an environment,
either because it is prohibitively expensive or dangerous, for example, in robotics or autonomous
driving~\cite{Prudencio2022OfflineRL}. Instead, we collect samples from expert demonstrations and
store them in a static dataset as trajectories prior to learning. In the simplest form, each
trajectory contains a sequence of state-action pairs.

Behaviour cloning aims to copy the target behaviour using supervised
learning~\cite{Prudencio2022OfflineRL}. We treat each state as input and each action as either a
class label in the case of discrete actions or as a real number in the case of continuous actions.
We learn the new policy by minimizing the appropriate loss function between the target's policy
actions and the new policy's outputs.

\subsection{Legends of Code and Magic}
\label{subsection:legends-of-code-and-magic}

Legends of Code and Magic (LOCM)~\cite{Kowalski2023SummarizingLOCM} is a two-player collectible
card game created to promote AI research. Like many CCGs, it consists of two stages: a draft stage
and a battle stage. In the draft stage, both players build their decks from a pool of available
cards. Once both players have built their decks, the game proceeds with the battle stage, in which
the players use their decks to decrease the opponent's health to zero. Unlike some other CCGs, card
effects in LOCM are completely deterministic with no random effects. The only non-determinism
during the game arises from the deck ordering at the beginning of the battle stage and partial
observability.

Two versions of the game are available online, version 1.2 and 1.5. The versions have a couple of
differences, but the main difference lies in the draft stage. Version 1.2 uses a fixed pool of 160
cards, whereas version 1.5 procedurally generates a pool of 120 cards before each game. Both
versions split the draft stage into 30 rounds. In LOCM 1.2, a player is presented with three cards
in each round and selects a single card with no restrictions. In LOCM 1.5, all 120 cards are
presented together, and a player chooses a single card in each round, with the restriction that the
final deck may contain at most two copies of the same card.

Even though Legends of Code and Magic is considered a small CCG compared to Magic: The Gathering or
Hearthstone, it is by no means a small game. LOCM 1.2 contains 160 available cards and has
approximately $(160 \times 159 \times 158)^{30} \approx 1.33 \times 10^{198}$ available
decks~\cite{Kowalski2020ActiveGenes}. This number is significantly larger than the number of atoms
in the universe. For comparison, Hearthstone, as of March 2024, features roughly 4500 cards. Even
though the number of possible decks in LOCM 1.2 is tremendous, simple rule-based agents and
heuristics based on card ordering were enough to reduce the game only to the battle stage. LOCM 1.5
and its procedural generation of cards changed that. The number of possible decks is even larger
and practically infinite, and the agents now must learn to generalize and deal even with unbalanced
cards (e.g. lethal cards with zero cost).

Our work currently focuses on version 1.5 of the game, so we shall describe what a typical game in
this version looks like. The game starts with the draft stage and proceeds according to the rules
above. In each timestep, each player receives an observation vector 2040 numbers long and chooses a
single card from the pool according to the rules by returning a number between 0 and 119. After the
30 timesteps, the draft stage concludes, and the game progresses to the battle stage. The player's
decks are randomly shuffled. The first player is dealt four cards, and the second player is dealt
five cards from their respective decks. The players take turns for up to 50 rounds or until one of
the player's health decreases to zero. Each player receives an observation vector 244 numbers long
and can perform multiple actions in each round, followed by a pass action to conclude their turn.
Actions are encoded by numbers 0 through 144. Every move receives an immediate reward of 0, except
for the last move, which receives either +1 when a player wins the game or -1 when a player loses
the game.

The original Java implementation of the game and reimplementations in Rust and Nim can be found
online~\cite{LocmGithub}. Additionally, the game has been rewritten in Python as an OpenAI Gym
environment~\cite{LocmGym}.

%%%%%%%%%%%%%%%%%%%%%%%%%%%%%%%%%%%%%%%%%%%%%%%%%%%%%%%%%%%%%%%%%%%%%%%%%%%%%%%%%%%%%%%%%%%%%%%%%%%

\section{Related Work}
\label{section:related-work}

In this section, we discuss prior work that has been done on collectible card games, mainly on
Legends of Code and Magic and Hearthstone. We divide this section into three subsections. The first
subsection summarises the Strategy Card Game AI Competition, which used Legends of Code and Magic
and was organized by the authors of the game. The second subsection focuses on research done in
Legends of Code and Magic, and Hearthstone. The last subsection is dedicated to the
state-of-the-art agent in Legends of Code and Magic~\cite{Xi2023MasteringLOCM}, and
Hearthstone~\cite{Xiao2023MasteringHS}.

\subsection{Strategy Card Game AI Competition}
\label{subsection:strategy-card-game-ai-competition}

Legends of Code and Magic initially started as an online challenge on the CodinGame platform in
2018. The first two online competitions attracted hundreds of players. The creators of Legends of
Code and Magic also organized a series of competitions between 2019 and 2022 named Strategy Card
Game AI Competition, which took place at the COG and CEC conferences organized by IEEE. They have
written a comprehensive paper detailing everything there is to know about LOCM, including
descriptions of different versions of the game and descriptions of every COG and CEC
participant~\cite{Kowalski2023SummarizingLOCM}. They have also publicly released the source codes
that were submitted to the competitions on their GitHub~\cite{LocmGithub}.

We shall briefly highlight the most frequent and the most important approaches that were submitted
to the competitions. As most of the participants did not write a paper detailing their approach, we
used the LOCM summary paper~\cite{Kowalski2023SummarizingLOCM} and the submitted code to figure out
the details.

During the earlier versions, including LOCM 1.2, participants mainly used handcrafted rules, such
as fixed deck ordering for the draft stage or lethal move detection during the battle stage.
Winning participants used different search techniques, such as Minimax with depth-limited search
and/or alpha-beta pruning or Monte Carlo Tree Search~\cite{Kowalski2023SummarizingLOCM}. Starting
at CEC 2020, the first neural network-based agent appeared in the competition. The agent used two
neural networks during the draft stage, one for each side trained by self-play. During the battle
stage, the agent used a best-first search with handcrafted rules optimized using Bayesian
optimization~\cite{Kowalski2023SummarizingLOCM}. Another notable approach that won COG 2021 was
using a flat simulation-based algorithm.

COG 2022 was the only competition to use the newer version of LOCM -- 1.5. This new version brought
a completely redesigned draft stage, rendering fixed deck ordering approaches unusable. Neural
network-based submissions dominated this competition. The approaches used mainly reinforcement
learning agents utilizing two-stage training where an agent is trained separately on each stage.
Some used Q-learning combined with best-first search, while others used PPO. ByteRL was the only
agent that used end-to-end training and won this competition by a large margin. We describe ByteRL
in more depth in Section~\ref{subsection:byterl}.

\subsection{Collectible Card Games}
\label{subsection:collectible-card-games}

Although most of the work on Legends of Code and Magic was done as part of the SCGAI Competition,
there have also been numerous papers on LOCM. Kowalski et al.~\cite{Kowalski2020ActiveGenes}
explored a variant of the evolutionary algorithm that uses a concept of active genes in the draft
stage of LOCM 1.2. Miernik et al.~\cite{Miernik2021EvolvingEvals} looked at evolving evaluation
functions for CCGs using genetic algorithms and genetic programming techniques. Yang et
al.~\cite{Yang2021DeckBuilding} also evolved various scoring functions using genetic algorithms,
compared the results against existing agents and analyzed the resulting decks in LOCM 1.2. Vieira
et al. published two papers that used self-play reinforcement learning to train agents for the
draft stage~\cite{Vieira2022ExploringRLDraft} and for the battle
stage~\cite{Vieira2022ExploringRLBattle} in LOCM 1.2. Vieira et
al.~\cite{Vieira2023TowardsSampleEff} also explored self-play reinforcement learning, reward
shaping and augmentation of features by adding information about the deck to the state
representation in LOCM 1.5.

Some research has also been conducted directly on Hearthstone. The Hearthstone AI
Competition~\cite{Dockhorn2019IntroducingHSComp} took place between 2018 and 2020. It featured two
tracks, both of which focused on the battle stage: 1) "pre-made deck" playing, which used three
publicly known decks and three unknown decks, and 2) "self-made deck" playing in which players had
to come up with their own deck ahead of time. Most of the submissions again used various search
techniques, evolutionary algorithms, heuristics or handcrafted rules. Janusz et
al.~\cite{Janusz2017HelpingAI} organized the AAIA'17 Data Mining Challenge to develop a state
evaluation function for Hearthstone. The goal was to predict the probability of winning given only
a single game state generated from Hearthstone matches between two random players. Hoover et
al.~\cite{Hoover2019TheHSChallenges} outlined many AI challenges Hearthstone poses, such as playing
the game to win, playing in a specific style, helping beginners to learn, helping professionals to
improve, balancing the decks and more. Bhatt et al.~\cite{Bhatt2018ExploringHS} explored the
Hearthstone deck space with evolutionary strategies by focusing on the deck-building phase.
Fontaine et al.~\cite{Fontaine2019MappingHSDecks} used an evolutionary approach in the form of a
modified MAP-Elites algorithm to create and balance decks in Hearthstone. Zhang et
al.~\cite{Zhang2021DeepSurrogate} extended MAP-Elites further using deep surrogate modelling and
applied it to deck-building in Hearthstone. Silva et al.~\cite{Silva2018HearthBot} used fuzzy ART
adaptive neural networks and trained an agent for the battle stage in Hearthstone. Xia et
al.~\cite{Xia2023Cardsformer} combined language modelling and self-play reinforcement learning to
train a battle-stage agent capable of outperforming the winning agent from Hearthstone AI
Competition 2020. Finally, Sakurai et al.~\cite{Sakurai2023RHEA} took a different approach to
beating some of the agents from the Hearthstone AI Competition using a modified version of the
Rolling Horizon Evolutionary Algorithm.

\subsection{ByteRL}
\label{subsection:byterl}

ByteRL~\cite{Xi2023MasteringLOCM} is the state-of-the-art agent submitted to the last LOCM
competition held in 2022. It is the first agent that views both stages in a unified manner. It is
trained end-to-end, meaning that a single trajectory contains states from both the draft stage and
the battle stage. The authors proposed Optimistic Smooth Fictitious Self-play (OSFP) to train
ByteRL, and unlike other agents capable of playing imperfect information games, it does not use
search. Agents using online search algorithms enumerate the world state consistent with the current
agent's information state. The number of such states in collectible card games is not tractable,
and any algorithm that explicitly needs to work with all the states is not feasible.

Optimistic Smooth Fictitious Self-play is an iterative algorithm using Smooth Best
Response~\cite{Fudenberg1998TheoryOfLearning} and reinforcement learning algorithms as building
blocks to find Nash Equilibria. It is an extension of the Smooth Fictitious Self-play
algorithm~\cite{Mertikopoulos2016Learning}. While in Smooth FSP the actual sequence of policies
does not converge to a Nash Equilibrium (only the sequence of average policies does), the authors
claim that in OSFP the actual sequence of policies converges to a Nash Equilibrium. For a more
in-depth explanation, we refer the reader to their paper~\cite{Xi2023MasteringLOCM}. To find the
smooth best response, ByteRL uses a policy gradient reinforcement learning algorithm as a
sub-solver. Specifically, they use the V-trace~\cite{Espeholt2018IMPALA} algorithm with an
auxiliary UPGO loss~\cite{Vinyals2019StarCraft}.

ByteRL uses a fairly complex neural network in the background. During the draft stage, all
available cards are embedded using a one-dimensional convolution layer, concatenated with already
selected cards and passed through another one-dimensional convolution layer. Subsequently, action
masking is applied and passed to the output head. For the battle stage, various features, such as a
player's hand, a player's deck, all cards on the table and other public information, are again
embedded using one-dimensional convolutions, concatenated and passed through three fully-connected
layers. Afterwards, a single LSTM layer combines the current output of the fully-connected layers
with the hidden state carried through the whole battle stage. Finally, as in the draft stage,
action masking is applied and passed to the output head.

%%%%%%%%%%%%%%%%%%%%%%%%%%%%%%%%%%%%%%%%%%%%%%%%%%%%%%%%%%%%%%%%%%%%%%%%%%%%%%%%%%%%%%%%%%%%%%%%%%%

\section{Behaviour Cloning}
\label{section:behaviour-cloning}

\begin{figure*}
    \centering
    \includegraphics[width=\textwidth]{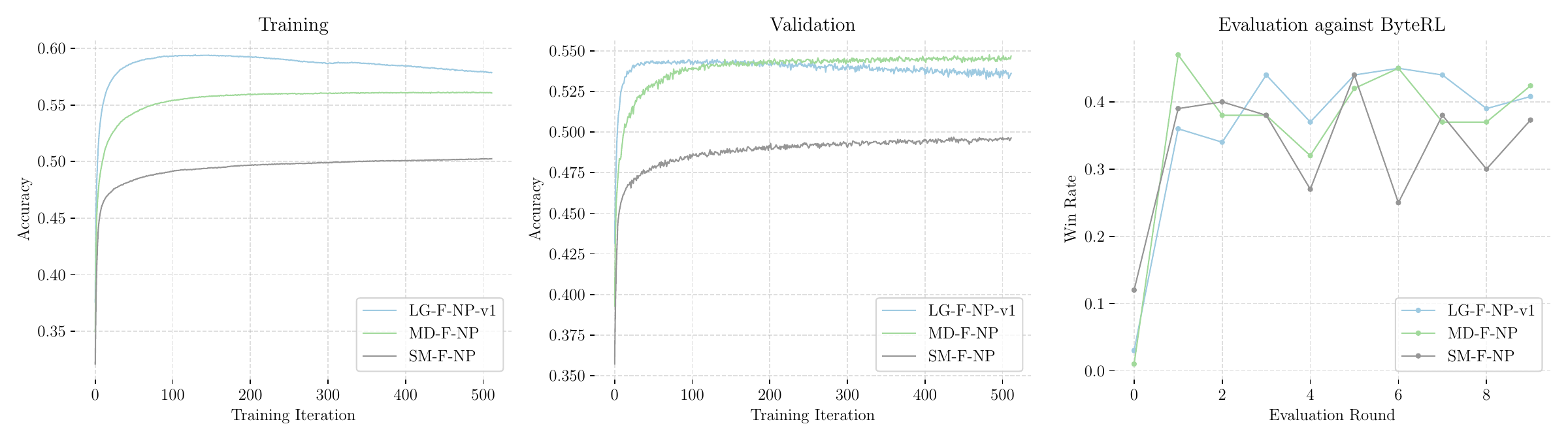}
    \caption{%
        The figure shows training, validation and win rate curves for selected behaviour cloning
        experiments. For the first two plots, the horizontal axis shows the training iteration, and
        the vertical axis shows the accuracy. For the last plot, the horizontal axis shows the
        evaluation round, and the vertical axis shows the win rate against ByteRL.
    }
    \Description{%
        The figure shows training, validation and win rate curves for selected behaviour cloning
        experiments. For the first two plots, the horizontal axis shows the training iteration, and
        the vertical axis shows the accuracy. For the last plot, the horizontal axis shows the
        evaluation round, and the vertical axis shows the win rate against ByteRL.
    }
    \label{figure:behaviour-cloning-accuracies}
\end{figure*}

\begin{figure}
    \centering
    \includegraphics[width=\columnwidth]{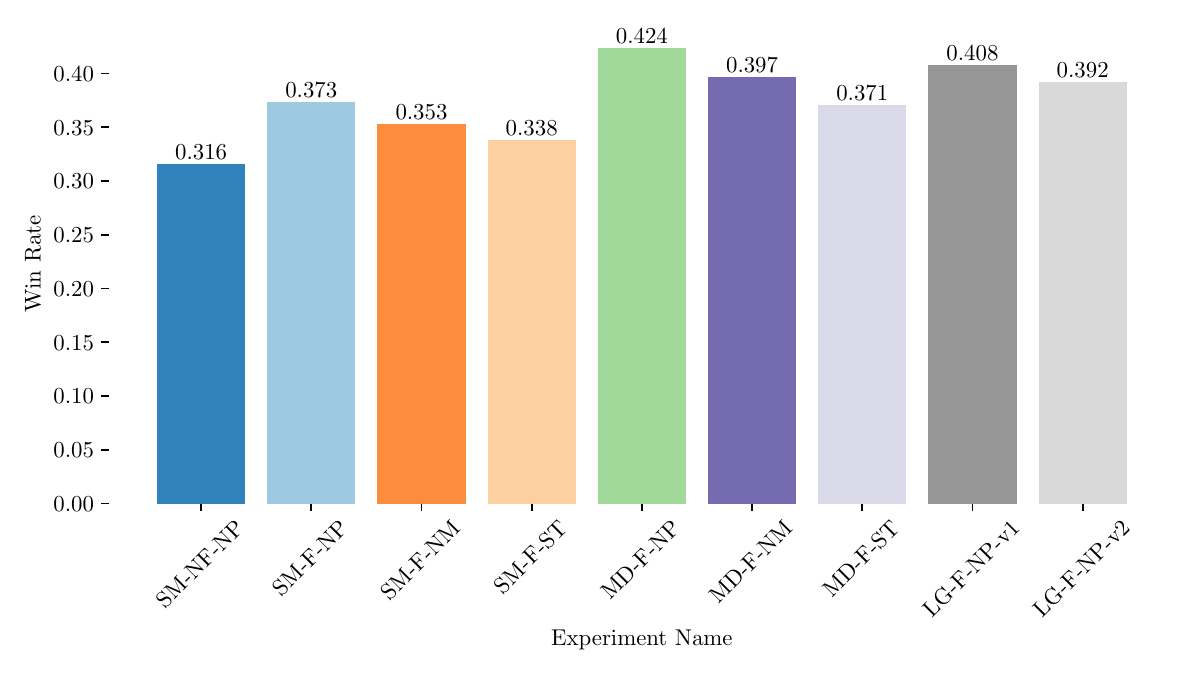}
    \caption{%
        The figure shows the results of the evaluations against ByteRL using the best-performing
        checkpoint for each behaviour-cloning experiment. The horizontal axis contains the names of
        the experiments, and the vertical axis shows the win rates.
    }
    \Description{%
        The figure shows the results of the evaluations against ByteRL using the best-performing
        checkpoint for each behaviour-cloning experiment. The horizontal axis contains the names of
        the experiments, and the vertical axis shows the win rates.
    }
    \label{figure:behaviour-cloning-winrates}
\end{figure}

In this section, we describe our experiments with behaviour cloning and present the results we
obtained. Although we could have used ByteRL with its existing weights that are available
online~\cite{LocmGithub} and fine-tuned it further, we chose to start from scratch and use
techniques that work even against black-box agents. It is important to note that although LOCM is a
two-stage game consisting of a draft stage and a battle stage, we have only focused on the battle
stage so far. The draft stage in LOCM 1.5 is not as straightforward to learn as the battle stage,
and it requires more engineering effort. Thus, whenever we evaluate an agent in the experiments, we
use ByteRL to perform the draft, providing us with a strong deck, and then we use the agent to
battle ByteRL with the drafted deck.

We started our experiments by using supervised learning to pre-train a policy network to mimic the
behaviour of ByteRL. We collected over 125 thousand matches between two instances of ByteRL playing
against each other. These matches produced roughly 3.5 million state-action pairs. We used 110
thousand matches (3 million state-action pairs) as our training data and the remaining 15 thousand
matches (500 thousand state-action pairs) as our validation data. The behaviour cloning experiments
were trained for 512 epochs using the same training and evaluation data. We used the Adam optimizer
with the default learning rate of 0.001 and a batch size of 256 to optimize the categorical
cross-entropy loss. We ran a periodic evaluation against ByteRL for 100 matches every 64 epochs. At
the end of every training, we restored the best-performing checkpoint and evaluated it against
ByteRL in 1000 matches.

\subsection{Filtering out the Pass Actions}
\label{subsection:filtering-out-the-pass-actions}

As every game contains many pass actions, either to end the turn or to select the first possible
action to play when in doubt, we trained two identical neural networks, with one neural network
using all of the data and the other using only non-pass actions. Both neural networks consisted of
one hidden layer with 128 neurons.

Filtering out the pass actions reduced the number of state-action pairs by roughly 25\%, from 3
million to 2.3 million in the training data and from 500 thousand to 330 thousand in the evaluation
data. The two experiments named \texttt{SM-NF-NP} and \texttt{SM-F-NP} in
Figure~\ref{figure:behaviour-cloning-winrates} show the win rates achieved in the 1000 evaluation
matches against ByteRL. Filtering out the pass actions increased the win rate by almost 7\%.
Figure~\ref{figure:behaviour-cloning-accuracies} shows the accuracy achieved on the training and
validation sets and how the evaluation results against ByteRL developed throughout training for the
experiment \texttt{SM-F-NP}.

As the environment automatically adds a pass action at the end of the list of actions returned by
an agent when needed, there is no need to keep training the future agents on datasets containing
pass actions. Thus, every subsequent experiment in this paper filters out the pass actions.

\subsection{Normalization \& Standardization}
\label{subsection:normalization-standardization}

The next batch of experiments looked at the preprocessing stage of the training pipeline. We tried
the following approaches: 1) using the raw input data without any preprocessing, 2) normalizing the
input data to the [0, 1] interval, and 3) standardizing the input data to have zero mean and unit
variance.

This time, we trained two architectures. The first one is the same as in the previous experiment: a
single hidden layer with 128 neurons. The second one is a scaled-up version of the first one. We
increased the number of hidden layers to two, and the layers contained 256 and 128 neurons,
respectively.

Looking at Figure~\ref{figure:behaviour-cloning-winrates}, experiment \texttt{SM-F-NM} represents
the smaller neural network with normalization, and \texttt{SM-F-ST} represents the same neural
network with standardization. Both neural networks achieved a lower win rate than the neural
network that uses raw input data (\texttt{SM-F-NP}).

The following three experiments are the larger networks with no processing (\texttt{MD-F-NP}),
normalization (\texttt{MD-F-NM}) and standardization (\texttt{MD-F-ST}), respectively. As we can
see, the results are the same as with the smaller network. No preprocessing achieved the best win
rate among the three networks, with normalization coming in second, slightly better than
standardization. Moreover, the larger network with no preprocessing achieved the best win rate
overall -- 42.4\%. It is important to note that we have achieved this win rate with no recurrence
and no memory. Figure~\ref{figure:behaviour-cloning-accuracies} shows the training and evaluation
curves for this network.

\subsection{Scaling up the Networks}
\label{subsection:scaling-up-the-networks}

Having decided on filtering out the pass actions and no preprocessing, we scaled the neural
networks in both the depth and the width even more. The results of these experiments are shown in
the remaining two bars. Making the network wider (\texttt{LG-F-NP-v1}) resulted in an inferior win
rate compared to our best-performing network, and we started noticing overfitting on the validation
data quite early in the training; see Figure~\ref{figure:behaviour-cloning-accuracies} for details.
We saw the same problem with the deeper network (\texttt{LF-F-NP-v2}), which consisted of three
layers with 256, 128 and 64 neurons, respectively.

The future plans in this area are the following. We have yet to see whether increasing the training
data size would prevent overfitting and allow training larger networks and whether it would
increase the win rate. We also plan to experiment with more complex architectures, such as
recurrent neural networks, optionally combined with one-dimensional convolutions. Lastly, we plan
to include a second head for value function prediction, which is currently trained from scratch
during reinforcement learning.

%%%%%%%%%%%%%%%%%%%%%%%%%%%%%%%%%%%%%%%%%%%%%%%%%%%%%%%%%%%%%%%%%%%%%%%%%%%%%%%%%%%%%%%%%%%%%%%%%%%

\section{Reinforcement Learning Fine-Tuning}
\label{section:rl-fine-tuning}

Although behaviour cloning yielded a stronger agent than we had initially expected, it still did
not match the performance of ByteRL. Thus, we started running another batch of experiments focusing
on reinforcement learning fine-tuning to learn the best response to ByteRL. Using reinforcement
learning is a common way to find an approximate best response to an agent in large games, where the
explicit computation of the best response is not tractable~\cite{Greenwald2017BestResponse,
Timbers2020ApproximateExploit}. In these experiments, we start with our best pre-trained network
and fine-tune it with the help of reinforcement learning. To justify the pre-training phase, we
compare the agents trained starting from the pre-trained network and agents trained from scratch.

All of the experiments in this section use the Proximal Policy Optimization (PPO)
algorithm~\cite{Schulman2017PPO} from the RLlib library~\cite{Liang2017RLlib}. The policy network
is initialized with the weights of the network trained using behaviour cloning that achieved the
highest win rate (\texttt{MD-F-NP}). The value network's weights are initialized randomly, and the
two networks do not share any weights. We periodically evaluate our agents for 100 matches every 20
training iterations. During evaluation, instead of sampling the action to take, we always take the
one with the highest probability. Evaluation happens at least five times during training, and we
stop the training at either 1000 iterations or when the average evaluation win rate from the last
five evaluation rounds surpasses 75\%. We use the default hyperparameters for PPO, except for the
increased batch sizes, to better utilize the hardware and an added entropy regularizer with a
coefficient of 0.01.

\subsection{Training on Fixed Decks}
\label{subsection:training-on-fixed-decks}

\begin{figure*}
    \centering
    \includegraphics[width=\textwidth]{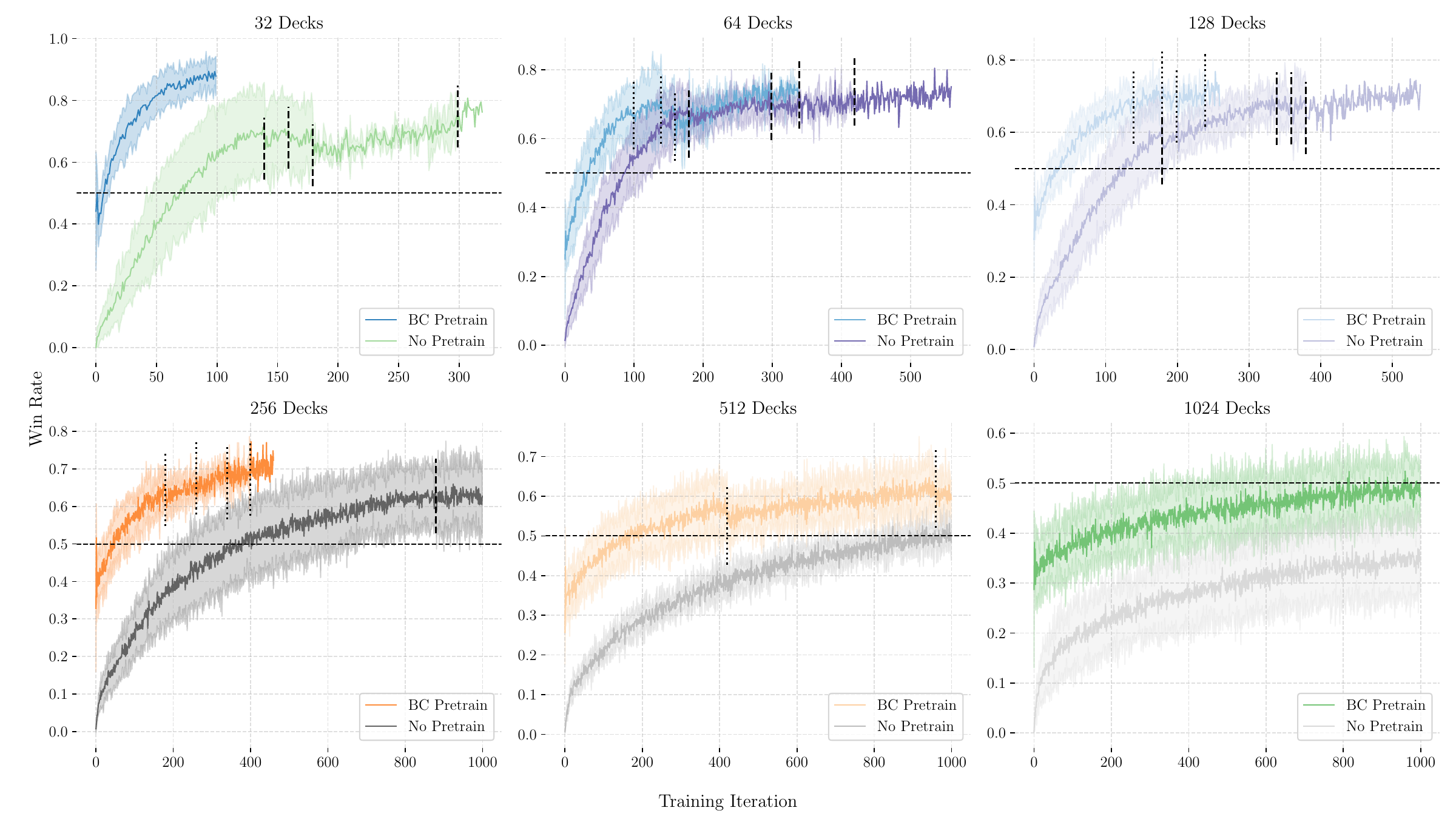}
    \caption{%
        The figure shows the development of win rates during the training. For each deck pool size,
        two curves with 95\% confidence intervals are shown, one for the case where the policy
        network was initialized with the pre-trained weights and one where the weights were
        initialized randomly. Each curve is an average of at most five runs. As the training on
        some of the seeds finished earlier than on others, the vertical bars denote the place where
        the number of runs decreased by one.
    }
    \Description{%
        The figure shows the development of win rates during the training. For each deck pool size,
        two curves with 95\% confidence intervals are shown, one for the case where the policy
        network was initialized with the pre-trained weights and one where the weights were
        initialized randomly. Each curve is an average of at most five runs. As the training on
        some of the seeds finished earlier than on others, the vertical bars denote the place where
        the number of runs decreased by one.
    }
    \label{figure:rl-training-winrates}
\end{figure*}

For our initial experiments, we decided to simplify the game by limiting the number of procedurally
generated deck pools. Fixing the number of unique deck pools allows us to control the game's
complexity and gradually increase the number of deck pools, reaching the full game at the end
without affecting the rules in any way. We ran several experiments with deck pool sizes of up to
1024 different deck pools.

To ablate the effect of the supervised pre-training, we ran two categories of experiments. The
policy network in the first category of experiments was initialized with the weights of the best
pre-trained network (\texttt{MD-F-NP}), while in the second category, the policy network was
randomly initialized. Figure~\ref{figure:rl-training-winrates} shows the averaged training curves,
and Table~\ref{table:rl-eval-winrates} shows the final evaluation results of these experiments.
Each combination of the initialization method and the deck pool size was run with five different
seeds, and the mean performance and 95\% confidence interval were computed. However, due to time
constraints, experiments whose average win rate over the last five evaluation rounds exceeded 75\%
were terminated, resulting in some runs having different lengths. In such cases, the averages and
confidence intervals are computed with the available data, and the decrease in the number of runs
being averaged is depicted using vertical lines. Dotted lines are used for experiments starting
from the pre-trained weights, and dashed lines are used for experiments starting with random
weights.

We were able to fine-tune the best pre-trained agent and beat ByteRL on up to 512 decks and match
its performance on 1024 decks during training, and we beat ByteRL in all six cases during
evaluation rounds. In the experiments on up to 256 decks, the win rate surpassed the 50\% mark in
less than 100 training episodes and reached the 75\% mark in less than 500 training episodes. The
fifth experiment needed a little more time, but all five runs eventually surpassed the 50\% mark,
and two of those reached the 75\% mark before the end of training. The last experiment matched
ByteRL's performance during training and slightly outperformed ByteRL during evaluation.

\begin{table}
    \centering

    \begin{tabular}{lrrrrrr}
        & 32 & 64 & 128 & 256 & 512 & 1024 \\
        \midrule
        BC Pretrain & 0.904 & 0.822 & 0.803 & 0.801 & 0.734 & 0.542 \\
        No Pretrain & 0.812 & 0.780 & 0.812 & 0.720 & 0.533 & 0.418 \\
    \end{tabular}

    \vspace{2mm}

    \caption{%
        The table shows the average win rate for each of the fixed-decks experiments. The average
        was computed from the last evaluation round for each of the five runs. The rows show the
        names of the two categories of experiments we ran and the columns show the number of deck
        pools available in the environment.
    }
    \Description{%
        The table shows the average win rate for each of the fixed-decks experiments. The average
        was computed from the last evaluation round for each of the five runs. The rows show the
        names of the two categories of experiments we ran and the columns show the number of deck
        pools available in the environment.
    }
    \label{table:rl-eval-winrates}

    \vspace{-6mm}
\end{table}

We repeated the same experiments with random initialization, plotted the results in
Figure~\ref{figure:rl-training-winrates} and showed the average evaluation win rates in
Table~\ref{table:rl-eval-winrates}. Apart from the second experiment, where the two curves overlap,
the rest of the experiments clearly showed that behaviour cloning prior to RL training is
beneficial. Although the first four experiments again defeated ByteRL, they required significantly
more time. From the first three experiments, we can see that the time needed to reach the win rate
of 75\% or more is more than twice that of experiments which started with the pre-trained weights.
The fourth experiment surpassed the 75\% win rate only in one out of five runs. The fifth
experiment got on par with ByteRL, and the final experiment did not even reach the win rate of our
pre-trained network (42.4\%). Finally, looking at Table~\ref{table:rl-eval-winrates}, we see that
with 512 and 1024 deck pools, the gap between the two categories of experiments grew significantly.

The next step in our experiments with reinforcement learning is to use curriculum learning to
automatically increase the number of available deck pools during training when the win rate reaches
a certain threshold. Furthermore, we plan to repeat the same experiments once we address the
shortcomings of the behaviour cloning pre-training as outlined in
Section~\ref{section:behaviour-cloning}.

%%%%%%%%%%%%%%%%%%%%%%%%%%%%%%%%%%%%%%%%%%%%%%%%%%%%%%%%%%%%%%%%%%%%%%%%%%%%%%%%%%%%%%%%%%%%%%%%%%%

\section{Conclusion}
\label{section:conclusion}

In this paper, we made the first steps in our work on collectible card games by focusing on one
simple collectible card game called Legends of Code and Magic. We put ByteRL, the state-of-the-art
agent in Legends of Code and Magic, to the test and saw that given a strong deck for the battle
stage, ByteRL's performance in the battle stage is highly exploitable, leaving space for further
improvement. We showed that simple behaviour cloning of ByteRL's policy yielded an agent that was
almost on par with ByteRL. Further fine-tuning led to an agent capable of matching or even
improving on ByteRL's performance on hundreds of decks. Lastly, we ran a small ablation study,
which showed that behaviour cloning prior to reinforcement learning fine-tuning is beneficial.
Although these results seem favourable, we are aware of the shortcomings of our preliminary
experiments, and we plan to continue working on them.

One of the most critical next steps is losing any dependence on ByteRL during the deck-building
stage, which requires training a separate network for the draft stage that can produce similarly
strong decks. Other steps will include continuing our work on both training phases, the behaviour
cloning phase and the fine-tuning phase. As mentioned in the paper, we plan to collect more
training data for the behaviour cloning phase, scale the networks even further, and see if that
increases the win rate against ByteRL. We also plan to experiment with different neural network
architectures, and lastly, we plan to train the value function during supervised training along the
policy network. In the RL fine-tuning phase, we plan to experiment with automatic curriculum
learning during the battle stage, where we automatically increase the number of deck pools during
training. We also plan to experiment with reinforcement learning during the draft stage.

%%%%%%%%%%%%%%%%%%%%%%%%%%%%%%%%%%%%%%%%%%%%%%%%%%%%%%%%%%%%%%%%%%%%%%%%%%%%%%%%%%%%%%%%%%%%%%%%%%%

\begin{acks}
    This publication was supported by Charles Univ. project UNCE 24/SCI/008.
    Computational resources were provided by the e-INFRA CZ project (ID:90254),
    supported by the Ministry of Education, Youth and Sports of the Czech Republic.
\end{acks}

%%%%%%%%%%%%%%%%%%%%%%%%%%%%%%%%%%%%%%%%%%%%%%%%%%%%%%%%%%%%%%%%%%%%%%%%%%%%%%%%%%%%%%%%%%%%%%%%%%%

\bibliographystyle{ACM-Reference-Format}
\bibliography{references}

\end{document}